\documentclass[a4paper,twocolumn]{article}
\usepackage{times}
\usepackage{helvet}
\usepackage{courier}
\usepackage[margin=2truecm]{geometry}
\usepackage{url}
\usepackage{graphicx}
\usepackage[breaklinks]{hyperref}
\usepackage[square,numbers]{natbib}
\usepackage{caption}
\usepackage{amsmath,amsfonts}
\usepackage{algorithm}
\usepackage{algorithmic}
\usepackage{array}
\usepackage{textcomp}
\usepackage{stfloats}
\usepackage{verbatim}
\usepackage{balance}
\usepackage{authblk}
\usepackage{microtype}
\usepackage{bm}

\usepackage{booktabs}
\usepackage{xspace}

\allowdisplaybreaks
\makeatletter
\DeclareRobustCommand\large{\@setfontsize\large\@xipt{13}}
\DeclareRobustCommand\Large{\@setfontsize\Large\@xiipt{14}}
\DeclareRobustCommand\LARGE{\@setfontsize\LARGE\@xivpt{16}}
\DeclareRobustCommand\huge{\@setfontsize\huge\@xviipt{20}}
\DeclareRobustCommand\Huge{\@setfontsize\Huge\@xxpt{23}}

\DeclareRobustCommand\onedot{\futurelet\@let@token\@onedot}
\def\@onedot{\ifx\@let@token.\else.\null\fi\xspace}

\def\eg{\emph{e.g}\onedot} 

\def\ie{\emph{i.e}\onedot}

\makeatother

\usepackage{pifont}
\newcommand{\cmark}{\ding{51}}%
\newcommand{\xmark}{\ding{55}}%

\usepackage[capitalize]{cleveref}
\crefname{section}{Sec.}{Secs.}
\Crefname{section}{Section}{Sections}
\Crefname{table}{Table}{Tables}
\crefname{table}{Tab.}{Tabs.}

\renewcommand{\vec}{\bm}

\newcommand{\R}{\mathbb{R}}

\newcommand{\vecp}{\vec{p}}
\newcommand{\vecq}{\vec{q}}
\newcommand{\vecy}{\vec{y}}
\newcommand{\vecz}{\vec{z}}
\newcommand{\veczero}{\vec{0}}

\newcommand{\thpred}{\theta_\mathrm{pred}}

\newcommand{\loss}{L}
\newcommand{\subrm}[2][]{_{\mathrm{#2}#1}}
\newcommand{\subpt}[1][]{\subrm[#1]{pt}}
\newcommand{\subvid}[1][]{\subrm[#1]{vid}}
\newcommand{\subpascl}[1][]{\subrm[#1]{pascl}}
\newcommand{\losspt}[1][]{\loss\subpt[#1]}
\newcommand{\lossvid}[1][]{\loss\subvid[#1]}
\newcommand{\losspascl}[1][]{\loss\subpascl[#1]}

\newcommand{\labels}{\mathcal{L}}
\newcommand{\videos}{\mathcal{V}}

\usepackage{comment}
\usepackage{multirow}

\begin{document}
\title{\textbf{Action-Agnostic Point-Level Supervision for Temporal Action Detection}}
\author[1]{Shuhei~M.~Yoshida}
\author[1]{Takashi~Shibata}
\author[1]{Makoto~Terao}
\author[2,3]{Takayuki~Okatani}
\author[3,4]{Masashi~Sugiyama}
\affil[1]{Visual Intelligence Research Laboratories, NEC Corporation, Kanagawa 211-8666, Japan}
\affil[2]{Graduate School of Information Sciences, Tohoku University, Miyagi 980-8579, Japan}
\affil[3]{RIKEN Center for Advanced Intelligence Project, Tokyo 103-0027, Japan}
\affil[4]{Graduate School of Frontier Sciences, The University of Tokyo, Chiba 277-8561, Japan}

\date{}

\maketitle

\begin{abstract}
We propose action-agnostic point-level (AAPL) supervision for temporal action detection to achieve
accurate action instance detection with a lightly annotated dataset. In the proposed scheme, a small
portion of video frames is sampled in an unsupervised manner and presented to human annotators, who
then label the frames with action categories. Unlike point-level supervision, which requires
annotators to search for every action instance in an untrimmed video, frames to annotate are
selected without human intervention in AAPL supervision. We also propose a detection model and
learning method to effectively utilize the AAPL labels. Extensive experiments on the variety of
datasets (THUMOS\,'14, FineAction, GTEA, BEOID, and ActivityNet 1.3) demonstrate that the proposed
approach is competitive with or outperforms prior methods for video-level and point-level
supervision in terms of the trade-off between the annotation cost and detection performance.\footnote{
An annotation tool to produce AAPL labels and the experimental code for training and testing are available at %
\url{https://github.com/smy-nec/AAPL}.}
\end{abstract}

\section{Introduction}
\label{sec:intro}

Temporal action detection is a vital research area in computer vision and machine
    learning, primarily focusing on recognizing and localizing human actions and events in
    untrimmed video
    sequences~\cite{Xia_TAD_Survey_Access_2020,Vahdani_TADSurvey_TPAMI_2022}.
With the rapid growth of video data available online, developing algorithms capable
    of understanding and interpreting such a wealth of information is critical for a
    wide range of applications, including anomalous event detection in surveillance
    videos~\cite{Vishwakarma_Surveillance_Survey_2013,Sultani_Anomaly_CVPR_2018} and
    sports activity analysis~\cite{Giancola_SoccerNet_2018,Cioppa_CVPR_2020}.
The existing literature generally tackles action detection problems through fully
    supervised approaches~\cite{Lin_BMN_ICCV_2019,Xu_GTAD_2020,Zhang_ActionFormer_2022},
    which require training data with complete action labels and their precise temporal
    boundaries.
Despite significant progress in recent years, these methods confront considerable
    challenges due to the high annotation cost to predict actions in complex and diverse
    video settings accurately.
To reduce the annotation cost for temporal action detection, weak supervision, such as
video-level
supervision~\cite{Sun_WTAL_2015,Wang_UntrimmedNets_2017,Baraka_WTAL_Survey_2022,Li_WTAL_Survey_2024}
and point-level labels~\cite{Moltisanti_Timestamp_2019,Ma_SF-Net_ECCV_2020}, has been
studied.
However, these types of supervision have their own difficulty in practice.
\begin{figure}[tb]
    \centering
    \includegraphics[width=0.9\columnwidth]{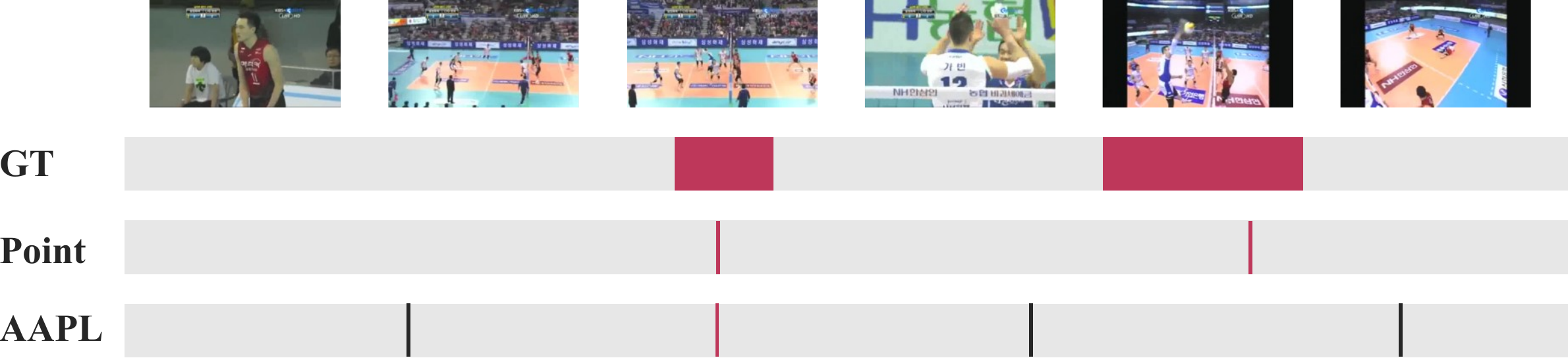}
    \caption{Illustration of ground-truth (full supervision), point-level supervision,
    and AAPL supervision. The red boxes and lines represents the frames labeled as
    ``Volleyball Spiking'', and the black lines represents those labeled as
    ``Background''. The images are from a video in
    THUMOS\,'14~\cite{Jiang_THUMOS14_2014}.}
    \label{fig:supervision}
\end{figure}

\begin{table*}[tb]
    \centering
    {
    \begin{tabular}{@{}lccccc@{}}
    \toprule
    & \multicolumn{4}{c}{Instance-level localization}
    & Video-level classes
    \\ \cmidrule(lr){2-5} \cmidrule(lr){6-6}
    & Available & Foreground & Background & Exhaustive
    & Exhaustive
    \\
    \midrule
    Full &
        \cmark & Complete & Complete & \cmark &
        \cmark 
        \\
    Video & 
        \xmark & --- & --- & --- &
        \cmark 
        \\
    Point &
        \cmark & Single point & \xmark & \cmark &
        \cmark 
        \\
    AAPL (ours) &
        \cmark & Point(s) & Point(s) & \xmark &
        \xmark 
        \\
    \bottomrule
    \end{tabular}
    }
    \caption{Comparison of full, video-level, point-level, and AAPL supervision.
    The first set of columns compares them from the perspective of four aspects of instance-level localization: whether information localizing action instances is available, what type of localization is given for foreground and background, and whether all the action instances are exhaustively annotated.
    The last column shows whether these types of supervision exhaustively give action classes appearing in each video.
    }
    \label{tab:supervisions}
\end{table*}

Video-level supervision only uses the action classes pre-sent in the video as
labels.
Various approaches have been proposed, such as multiple
instance learning-based~\cite{Wang_UntrimmedNets_2017,Paul_W-TALC_ECCV_2018}, feature
erasing-based~\cite{Singh_Hide-and-Seek_2017}, and
attention-based~\cite{Nguyen_STPN_2018,Lee_BASNet_2020,Liu_CMCS_2019}, but they
ultimately reduce action detection learning to video classification.
Behind this strategy is the assumption that the discriminative intervals contributing to
classification are where the actions occur, which is not always true.
In addition, when a video contains multiple action classes, it becomes a multi-label
classification problem, which is extremely difficult.
These limitations severely limit the range of applications of video-level supervision.

Point-level supervision specifies for each action instance a single arbitrary time point in the
instance and the action class and has been actively studied
recently~\cite{Ju_Point-Level_ICCV_2021,Lee_LACP_ICCV_2021,Li_PCL_ESWA_2023,Li_Point-Level_CVPR_2021}.
While the point-level labels convey partial information about the location of action
instances, they do not tell where the actions are \emph{not}.
This is a fundamental difficulty in point-level supervised learning of action detection
because localization is to distinguish actions from non-actions.
In addition, the requirement that action instances must be exhaustively labeled makes
the annotation process expensive.

To achieve better trade-off between the annotation costs and detection accuracy, we propose
action-agnostic point-level (AAPL) supervision, a novel form of weak supervision for temporal action
detection (\cref{fig:supervision}).
In producing AAPL labels, a small portion of video frames is sampled and presented to
    human annotators, who label the frames with action categories.
Unlike point-level supervision, where annotators need to find all action instances in
    untrimmed videos, the frames to annotate are selected without human intervention. 
We also propose a baseline learning protocol exploiting the AAPL labels.
Our proposed method, such as the point loss, the video loss, and the prototype-anchored
    supervised contrast loss, adapts ideas from previous studies for the current setting to
    exploit the AAPL labels.
To demonstrate the utility of AAPL supervision in various use cases,
we empicically evaluate our approach using five dataests with different characteristics, including
BEOID~\cite{Damen_BEOID_2014}, GTEA~\cite{Fathi_GTEA_2011}, THUMOS\,'14~\cite{Jiang_THUMOS14_2014},
FineAction~\cite{Liu_FineAction_TIP_2022}, and ActivityNet 1.3~\cite{Caba_ActivityNet_CVPR_2015}.
The results show that the proposed method is competitive with or outperforms prior methods for
video-level and point-level supervision.
We visualize the trade-offs between the costs and detection performance and compare AAPL
    supervision with other supervision types.
We also find that even training only with annotated frames can achieve competitive results with the
previous studies.
This suggests the inherent effectiveness of AAPL supervision.

The contributions of this paper are as follows:
\begin{itemize}
\item We propose AAPL supervision, a novel form of weak supervision for temporal action detection,
    which achieves good
    trade-offs between the annotation costs and detection accuracy.
\item We design an action detection model and loss functions that can leverage the
    AAPL-labeled datasets.
\item Comprehensive experiments on a wide range of action detection benchmarks demonstrate that the
    proposed approach is competitive with or outperforms previous methods using video-level or
    point-level supervision in terms of the trade-off between the annotation cost and detection
    performance.
\end{itemize}

\section{Action-Agnostic Point-Level Supervision}
\label{sec:aapls}

We first explain the annotation pipeline for AAPL supervision (\cref{sec:def-notations}).
Then, we compare AAPL supervision with other forms of weak supervision qualitatively
(\cref{sec:aapls-comparison}) and in terms of annotation time (\cref{sec:annotation-time}).
Some notations for AAPL labels are also introduced (\cref{sec:notations}).

\subsection{Annotation Pipeline}
\label{sec:def-notations}
AAPL supervision is characterized by the two-step annotation pipeline consisting of action-agnostic
frame sampling and manual annotation. Action-agnostic frame sampling determines which frames in the
training videos to annotate. This can be an arbitrary method that, without any human
intervention, selects video frames to annotate. Then, human annotators label the sampled frames with action
categories.

Action-agnostic frame sampling is what distinguishes AAPL supervision from conventional point-level
supervision. The previous scheme requires that every action instance in a video be annotated with a
single time point. This is a challenging task because human annotators need to search videos for
every action instance. By contrast, for AAPL supervision, annotators just annotate the sampled
frames with action categories, but they do not need to search for action instances.

The simplest examples of action-agnostic frame sampling are regularly spaced sampling and random
sampling. The former picks up frames at regular intervals, while the latter selects frames randomly.
A strength of these methods is that they are easy to implement, computationally light, and free from
any assumption on the videos. As we will see in \cref{sec:aafs}, the regular sampling is more preferable than the random one
because the latter can result in multiple frames in temporal proximity being selected, leading to
redundant annotations. We can also consider more sophisticated sampling strategies that take into
account the content of the video. For example, we can use a pre-trained feature extractor to
compute the feature representations of the frames and then cluster the frames based on the features.
The representative frames of each cluster can be selected for annotation.
See \cref{sec:aafs} for performance comparison.

Because sampling at regular intervals is computationally free, it might be suitable as an initial
choice.
In addition, it involves only one hyper-parameter, the interval length, which can be sensibly
determined by using prior knowledge about the dataset, \eg, the duration and frequency of action
instances.
If one has the computational resources, sophisticated methods like clustering-based sampling can be
a good alternative, because it can adapt to the dataset characteristics and potentially provide better
performance.

\subsection{Qualitative Comparison}
\label{sec:aapls-comparison}

Here, we contrast AAPL supervision with other types of supervision.
\Cref{tab:supervisions} compares four supervision schemes for temporal action detection: full, video-level, point-level, and AAPL supervision.

Information about instance-level localization is partially available in the AAPL labels, which do
not give the exact starting and ending times of action instances but do include timestamps on them.
This is similar to point-level supervision, but AAPL supervision can have multiple labels on a
single action instance, conveying more complete information about the action location.
It also has labels on background frames, which is crucial for temporal localization because
localizing an action entails finding the boundaries between the action and the background.
Conventional point-level supervision contains timestamps of foreground frames only, and previous
work resorts to a self-training strategy to mine background frames, assuming there is at least one
background frame between two point-level labels~\cite{Lee_LACP_ICCV_2021}.
This assumption is plausible but has a minor practical meaning for rare actions because in such
cases point-level labels are distributed so sparsely that two point-level labels cannot effectively
narrow down the location of the action boundaries.

Action-agnostic frame sampling is not guaranteed to find all the action instances in a video, and
some action instances might not have labels.
This is a potential weakness of AAPL supervision.
This is true even at the video level; action-agnostic frame sampling might miss all the instances of
an action class that is indeed present in the video.
This makes it challenging to apply popular methods such as a video-level loss function, which is
known to be effective both for video-level~\cite{Paul_W-TALC_ECCV_2018}
and point-level supervision~\cite{Ma_SF-Net_ECCV_2020,Lee_LACP_ICCV_2021}.
However, this problem can be mitigated by a simple modification to the video-level loss introduced
in \cref{sec:method-objective}.

\subsection{Measurement of Annotation Time}
\label{sec:annotation-time}

We measured the annotation time for full, video-level, point-level, and AAPL supervision, using a
modified version of the VGG Image Annotator (VIA)~\cite{Dutta_VIA_2016,Dutta_VIA_2019}.
For AAPL supervision, we sampled frames to annotate at regular intervals of 3, 5, 10, and 30
seconds.
We had eight workers annotate the videos in
BEOID~\cite{Damen_BEOID_2014}, GTEA~\cite{Fathi_GTEA_2011}, and THUMOS\,'14~\cite{Jiang_THUMOS14_2014}.
The detailed protocol of this measurement is given in \cref{apdx:measurement}.

\begin{table*}[tb]
    \centering
    {
    \begin{tabular}{@{}lccccccc@{}}
    \toprule
    &      &       &       & \multicolumn{4}{c}{AAPL} \\ \cmidrule(lr){5-8}
    & Full & Video & Point & 3\,sec. & 5\,sec. & 10\,sec. & 30\,sec. \\
    \midrule
    BEOID & 3.72 & 1.11 & 2.44 & 2.09 & 1.43 & 0.94 & 0.45 \\
    GTEA & 4.49 & 0.93 & 3.03 & 1.98 & 1.60 & 1.09 & 0.53 \\
    THUMOS\,'14 & 1.92 & 0.45 & 1.10 & 1.31 & 0.95 & 0.64 & 0.36 \\
    \bottomrule
    \end{tabular}
    }
    \caption{Annotation time relative to the video duration.
    ``Full'', ``Video'', and ``Point'' represent the full segment-level supervision,
    video-level supervision, and point-level supervision, respectively.
    ``$T$\,sec.'' stands for the intervals for AAPL supervision.}\label{tab:annotation-cost}
\end{table*}

\Cref{tab:annotation-cost} shows the measured annotation time relative to the duration of videos, \ie, 
    the minutes it took for one annotator to annotate a 1-minute video.
The previous methods (``Full'', ``Video'', and ``Point'') exhibit the expected ordering that
    full supervision costs the most, and that video-level supervision costs the least.
On the other hand, the annotation time for AAPL supervision varies with the intervals
and is well-approximated by a linear function of the number of labeled frames.
This modeling assumes that the annotation time per frame is constant,
    which is reasonable because 
    the annotation time per frame is dominated by the time to select the action category and
    is not sensitive to the number of frames to annotate.

The annotation time depends on the dataset's characteristics, such as the density (\ie, the number
per unit length of a video) of action instances and the number of action classes occurring in one
video.
Indeed, these numbers are much larger in BEOID and GTEA than in THUMOS\,'14.
As a result, annotating videos in BEOID and GTEA takes over twice as long as annotating those in
THUMOS\,'14 for full, video-level, and point-level supervision.
By contrast, the variation in the annotation time is relatively small for AAPL supervision because
AAPL annotation involves local segments around the frames to label and is insensitive to global
characteristics like density.
This property makes it easy to apply AAPL supervision in a variety of datasets.

\subsection{Notations}
\label{sec:notations}

We introduce notations for AAPL labels.
Let $\videos$ be a set of videos.
AAPL labels for a video $V \in \videos$ are a set $\labels^V=\{(t_i, \vecy_i)\}_{i\in[N^V]}$ of pairs of a time stamp $t_i$ and an action label $\vecy_i \in \{0,1\}^C$.
Here, $N^V$ is the number of annotated frames, $C$ is the number of action categories, an action label $\vecy_i$ is a 0/1-valued vector with the $c$-th component indicating the presence or absence of action of the $c$-th class at the time $t_i$, and $[K]$ is the set $\{1,2,\dots,K\}$.
An annotated frame might not belong to any action instance. Such a frame is called a background and labeled $\vecy = \veczero$.
Also, if multiple action instances of different categories overlap, the frames in the intersection are annotated with a multi-hot vector representing all the action categories present there.

\section{AAPL Supervised Learning Method}\label{sec:method}

This section explains our approach to temporal action detection under AAPL supervision.
This includes
    the action detection pipeline predicting action instances from an input video (\cref{sec:method-pipeline}), 
    the training objectives for the prediction model (\cref{sec:method-objective}),
    and the pseudo-labeling strategy to make more effective use of the training data (\cref{sec:method-pseudolabels}).

\subsection{Action Detection Model}\label{sec:method-pipeline}

\begin{figure*}[tb]
    \centering
    \includegraphics[width=0.8\textwidth]{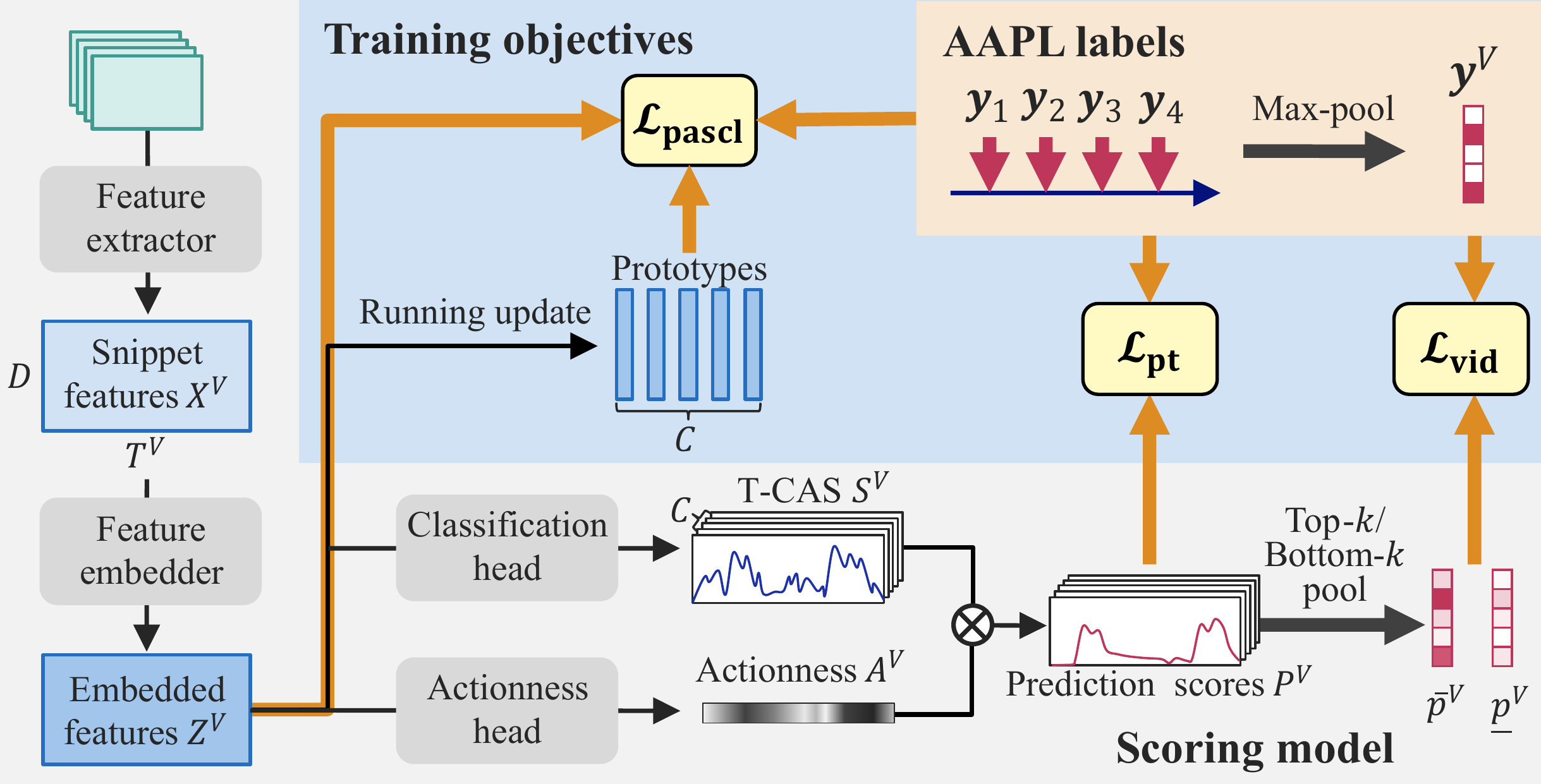}
    \caption{Illustration of the model and the loss functions.}
    \label{fig:method}
\end{figure*}

Our action detection pipeline consists of preprocessing, snippet scoring, and action instance
generation, following previous studies~\cite{Li_WTAL_Survey_2024},
At the preprocessing stage, we divide an input video into $T^V$ non-overlapping segments of $\tau$ frames called snippets and apply the transformation to make them fit the snippet scoring model.
The snippet scoring model processes an input video $V$ into a prediction score sequence $P^V \in \R^{C \times T^V}$.
The prediction score $P_{ct}^V$ indicates the likelihood of an action of the $c$-th class occurring at time $t$.
The action instance generator then converts the score sequence into a set of the scored action instances $\{(s_i, e_i, c_i, p_i)\}_{i\in[M^V]}$ in the video. Here, $M^V$ is the number of action predictions in the video, $s_i$ is the starting time, $e_i$ is the ending time, $c_i$ is the action category, and $p_i$ is the confidence score of the $i$-th prediction.

The snippet scoring model comprises a feature extractor, a feature embedder, and two scoring heads, as illustrated in \cref{fig:method}.
The feature extractor is a pretrained 3D CNN converting each preprocessed snippet into a $D$-dimensional snippet feature.
We denote the sequence of the snippet features by $X^V \in \R^{D\times T^V}$.
The feature sequence is further fed into the feature embedder, a temporal convolution layer of the kernel size three followed by the rectified linear unit activation, which outputs the embedded feature sequence $Z^V = (\vecz_1^V, \dots, \vecz_T^V) \in \R^{D\times T^V}$.
The embedded features are split into $Z^V_S \in \R^{D/2\times T^V}$ and  $Z^V_A \in \R^{D/2\times T^V}$, which are then input to the two scoring heads.
The classification head classifies each snippet and outputs class-specific classification scores
$S^V \in \R^{C\times T^V}$ called the temporal class activation sequence (T-CAS).
The actionness head calculates class-agnostic scores called the actionness sequence $A^V \in \R^{T^V}$, which represents the likelihood of a snippet being in an action instance.
Both heads combine a point-wise temporal convolution layer and the sigmoid function.
We chose the sigmoid instead of the softmax function as the activation for the T-CAS because multiple action instances of different classes might overlap, in which case the model must predict all the classes present at each moment as positive.
The final prediction scores are the product of the two score sequences: $P_{ct}^V = A_t^V S_{ct}^V$.

Given the prediction scores for an action category, the action instance generator first upsamples the score sequence to match the frame rate of the input video. 
It then generates a set of action candidates by collecting the intervals over which the prediction scores are above a threshold $\thpred$.
This process is repeated with several different thresholds.
Then, for each action candidate, it calculates the outer-inner contrastive score~\cite{Shou_AutoLoc-OIC_ECCV_2018} as the initial confidence score.
Finally, soft non-maximum suppression~\cite{Bodla_SoftNMS_ICCV_2017} removes duplicate predictions and we calculate the final confidence scores.

\subsection{Training Objectives for the Scoring Model}
\label{sec:method-objective}

Our training objective is the weighted sum of three terms:
\begin{align}
    \loss = \losspt + \lambda\subvid \lossvid + \lambda\subpascl \losspascl,
\end{align}
where $\losspt$ is the point-level classification loss, $\lossvid$ is the video-level classification loss, and $\losspascl$ is the prototype-anchored supervised contrastive loss (see \cref{fig:method}).
For brevity, we also call them the point loss, the video loss, and the contrastive loss, respectively.
Unless otherwise stated, the averaging over a mini-batch of videos is implied in the expressions of the loss functions below.

The \textbf{point-level classification loss} quantifies the classification error on labeled
snippets. We adopt the focal loss~\cite{Lin_FocalLoss_2020} for this purpose.
We separate the contributions from the foreground and background snippets,
$\losspt = \losspt[,\mathrm{fg}] + \losspt[,\mathrm{bg}]$, to handle the class imbalance between them:
\begin{align}
&\begin{aligned}
    \losspt[,\mathrm{fg}]
    &= \frac{-1}{|\labels^V_\mathrm{fg}|} \sum_{(t, \vecy) \in \labels^V_\mathrm{fg}} \Biggl\{
        (1-A_{t}^V)^2 \log A_{t}^V
     \\
    &\quad + 
    \sum_{c=1}^C \left[
        y_{c} (1-S_{ct}^V)^2 \log S_{ct}^V \right.\\
    &\quad \left.+ (1-y_{c}) (S_{ct}^V)^2 \log (1-S_{ct}^V)
    \right] 
    \Biggr\},
\end{aligned} \\
&\begin{aligned}
    \losspt[,\mathrm{bg}]
    &= \frac{-1}{|\labels_\mathrm{bg}^v|} \sum_{(t, \vecy) \in \labels_\mathrm{bg}^v}\biggl[
        (A_{t}^V)^2 \log (1-A_{t}^V)
     \\
    & \quad+
        \sum_{c=1}^c (s_{ct}^v)^2 \log (1-s_{ct}^v)
    \biggr].
\end{aligned}
\end{align}
Here, $\labels_\mathrm{fg}^V = \{ (t, \vecy) \in \labels^V \mid \vecy \neq \veczero \}$, and $\labels_\mathrm{bg}^V = \{ (t, \vecy) \in \labels^V \mid \vecy = \veczero \}$,
representing the subsets of AAPL labels on the foreground and background snippets, respectively.
Importantly, we can calculate both $\losspt[,\mathrm{fb}]$ and $\losspt[,\mathrm{bg}]$ by using human-generated AAPL labels only because AAPL labels have labels on background snippets.
By contrast, previous point-level methods~\cite{Ma_SF-Net_ECCV_2020,Lee_LACP_ICCV_2021} require pseudo-labeling to calculate the background point loss.
This is a significant advantage of AAPL supervision over previous point-level methods because action localization involves distinguishing foreground actions from the background and having reliable labels on the background snippets is crucial for learning this task.

The \textbf{video-level classification loss}
measures the agreement between the video-level labels and predictions.
This type of loss functions has played a central role in both video-level supervision~\cite{Wang_UntrimmedNets_2017,Paul_W-TALC_ECCV_2018} and
It has also been adopted for 
point-level supervision~\cite{Ma_SF-Net_ECCV_2020,Lee_LACP_ICCV_2021}.
These settings have complete video-level labels, \ie, the presence or absence of each action class in each video is known.
By contrast, in the AAPL-supervised setting the video-level labels might be incomplete.
In other words, the absence of AAPL labels of an action class does not necessarily imply that the class is absent in the video.
Consequently, we cannot simply apply the video loss as used in the video-level~\cite{Wang_UntrimmedNets_2017,Paul_W-TALC_ECCV_2018} and point-level~\cite{Ma_SF-Net_ECCV_2020,Lee_LACP_ICCV_2021} scenarios.

To handle this incompleteness, we introduce the positive and negative parts of the video loss.
The positive part of the video loss is expressed as%
\begin{align}
    \lossvid[,\mathrm{pos}] =
    -\sum_{c\in[C]} 
        y^V_c \log \bar{p}^V_c,
\end{align}
where $\bar{p}^V_c$ is the video-level prediction score,
\begin{align}
    \bar{p}^V_c = \sigma \left(
        \frac{1}{k_\mathrm{pos}} \max_{\mathcal{T} \subset [T^V], |\mathcal{T}| = k_\mathrm{pos}} \sum_{t\in\mathcal{T}}
        \sigma^{-1} \left( P_{ct}^V \right)
    \right),
\end{align}
$y^V_c$ is the video-level label,
\begin{align}
    y^V_c = \max_{(t, \vecy) \in \labels^V} y_c,
\end{align}
and $\sigma$ and $\sigma^{-1}$ represent the sigmoid function and its inverse function.
The terms of the form $(1-y^V_c) \log (1-\bar{p}^V_c)$ are excluded in our loss because $y^V_c=0$ does not
necessarily mean that the $c$-th class is absent in the video.
This exclusion can lead to a biased estimation of the video-level prediction scores.
To compensate this bias, we introduce the negative part of the video loss, $\lossvid[,\mathrm{neg}]$.
To this end, we define $\underline{\vecp}^V$ by the ``bottom-$k$'' pooling:
\begin{align}
    \underline{p}^V_c = \sigma \left(
        \frac{1}{k_\mathrm{neg}} \min_{\mathcal{T} \subset [T^V], |\mathcal{T}| = k_\mathrm{neg}} \sum_{t\in\mathcal{T}}
        \sigma^{-1} \left( P_{ct}^V \right)
    \right).
\end{align}
This represents the average scores of frames that are not likely in the $c$-th class.
Then, the negative part is written as
\begin{align}
    \lossvid[,\mathrm{neg}]
    = - \sum_{c\in[C]} \log \left( 1 - \underline{p}_c^V \right).
\end{align}
The total video loss is the simple sum of the two parts: $\lossvid = \lossvid[,\mathrm{pos}] + \lossvid[,\mathrm{neg}]$.

Following recent work~\cite{huangRelationalPrototypicalNetwork2020,Liu_CASE_ICCV_2023,Lee_LACP_ICCV_2021,Li_PCL_ESWA_2023} demonstrating that enhancing the discriminative power of embedded features improves action detection performance,
we also introduce the \textbf{prototype-anchored supervised contrastive loss}.
This loss was inspired by the SupCon loss~\cite{Khosla_SupCon_NeurIPS_2020} and utilizes AAPL labels to enhance the embedded features.
The anchors in the SupCon loss are replaced with the prototypes.
This modification makes our loss computationally more efficient.

To formulate the prototype-anchored supervised contra-stive loss,
we first introduce the prototype $\vecq_c$ for the $c$-th action class.
The prototype is the running estimate of the average embedded features for snippets belonging to action instances of the $c$-th class.
The prototype $\vecq_c$ is initialized as%
\begin{align}
    \vecq_c = \frac{1}{|\labels_c|} \sum_{(V, t, \vecy)\in \labels_c} \vecz_t^V,
\end{align}
where $\labels_c = \{(V, t,\vecy) \mid \forall V\in\videos, (t, \vecy) \in \labels^V, y_c= 1 \}$ is the set of all the AAPL labels attached to $c$-th class action snippets in the training dataset.
During the training, $\vecq_c$ is updated at every iteration as
\begin{align}
    \vecq_c \leftarrow (1-\mu) \vecq_c + \frac{\mu}{|L_c^\mathcal{B}|} \sum_{(V, t, \vecy)\in \labels_c^\mathcal{B}} \vecz_t^V,
\end{align}
where $\labels_c^\mathcal{B}$ is a subset of $\labels_c$ from the videos in the mini-batch for that iteration.

Using the prototypes, the prototype-anchored supervised contrastive loss for a mini-batch $\mathcal{B}$ is expressed as%
\begin{equation}
\begin{aligned}
    \losspascl
    &= \sum_{c\in[C]} \frac{-1}{|\labels_c^\mathcal{B}|} \\
    &\quad\times\sum_{(V, t, \vecy)\in \labels_c^\mathcal{B}}
    \log \left[
        \frac{e^{\vecq_c \cdot \vecz_t^V / \tau}}{\sum_{(V', t', \vecy') \in \labels^\mathcal{B}} e^{\vecq_c \cdot \vecz_{t'}^{V'} / \tau}}
    \right],
\end{aligned}
\end{equation}
where $\labels^\mathcal{B} = \cup_{c\in[C]} \labels_c^\mathcal{B}$.
Because $\losspascl$ is calculated using all the videos in the mini-batch, we do not apply mini-batch averaging to $\losspascl$.
This loss function pulls the embedded features of the $c$-th action class to $\vecq_c$ while repelling the others from it.
We do not use a prototype for background features, and therefore, such features are repelled by all the prototype vectors.

\subsection{Ground-Truth Anchored Pseudo-Labeling}
\label{sec:method-pseudolabels}

Among the loss functions in the previous section, the point loss $\losspt$ and the contrastive loss
$\losspascl$ do not involve unlabeled snippets, which constitute the majority of the snippets in the
training dataset.
Pseudo-labeling offers a convenient way of exploiting these underutilized data by generating
pseudo labels from the predictions and using them in calculating the losses.
To obtain a better outcome, the quality of the pseudo labels is crucial.

Here, we adopt the ground-truth anchored pseudo-label-ing strategy, inspired by \citet{Ma_SF-Net_ECCV_2020} and \citet{Li_PCL_ESWA_2023}.
Under this strategy, pseudo-labels of the $c$-th action class are assigned to the snippets on an
interval if
(i) the prediction scores $P_c^V$ over the interval are above a threshold $\theta_\mathrm{fg}$,
(ii) at least one of the snippets is annotated with an AAPL label, and 
(iii) every AAPL label $(t, \vecy)$ on the interval satisfies $y_c = 1$.
Put differently, pseudo-labels are given to intervals with highly confident predictions consistent with AAPL labels.
Similarly, pseudo-background labels are assigned to an interval if
(a) the actionness scores are below a threshold $\theta_\mathrm{bg}$ over the interval, and
(b) there is at least one background label and no foreground action label on the interval.
When calculating the point loss and the contrastive loss, we replace the AAPL labels with the pseudo labels.

\section{Experiments}
\label{sec:exp}

\begin{table*}[tb]
    \centering
    {
    \begin{tabular}{@{}llcccccccccc@{}}
    \toprule
    & & \multicolumn{5}{c}{mAP@IoU\,[\%] (BEOID)} & \multicolumn{5}{c}{mAP@IoU\,[\%] (GTEA)} \\ \cmidrule(lr){3-7} \cmidrule(lr){8-12}
    Supervision & Method &
        0.1 & 0.3 & 0.5 & 0.7 & Avg &
        0.1 & 0.3 & 0.5 & 0.7 & Avg \\ \midrule
    Point
        & \citet{Ma_SF-Net_ECCV_2020}
            & 62.9 & 40.9 & 16.7 & 3.5 & 30.9
            & 58.0 & 37.9 & 19.3 & 11.9 & 31.0
        \\
        & \citet{Ju_Point-Level_ICCV_2021}
            & 63.2 & 46.8 & 20.9 & 5.8 & 34.9
            & 59.7 & 48.3 & 21.9 & 18.1 & 33.7
        \\
        & \citet{Li_Point-Level_CVPR_2021}
            & 71.5 & 40.3 & 20.3 & 5.5 & 34.4
            & 60.2 & 44.7 & 28.8 & 12.2 & 36.4
        \\
        & \citet{Lee_LACP_ICCV_2021}
            & 76.9 & 61.4 & 42.7 & 25.1 & 51.8
            & 63.9 & 55.7 & 33.9 & 20.8 & 43.5
        \\
        & \citet{Li_PCL_ESWA_2023} 
            & \textbf{78.7} & 63.3 & 44.1 & \textbf{26.9} & 53.3
            & 65.2 & \textbf{56.8} & 34.3 & 21.2 & 44.9
        \\
    \midrule
    AAPL
    (3 sec.)    & \emph{Ours}
            & 75.5 & \textbf{67.6} & \textbf{48.5} & 26.3 & \textbf{55.2}
            & \textbf{70.3} & 54.4 & \textbf{37.7} & \textbf{23.4} & \textbf{46.3}
        \\
    \bottomrule
    \end{tabular}
    }
    \caption{Detection performance on GTEA and BEOID. Each column shows the mAP
    at a specific IoU threshold (0.1, 0.3, 0.5, and 0.7) and the average mAP (Avg) over the thresholds.}
    \label{tab:gtea-beoid}
\end{table*}

In this section, we empirically evaluate the effectiveness of AAPL supervision for temporal action
detection.
As action-agnostic frame sampling, we use the regularly spaced sampling, except in the part of
\cref{sec:aafs} that compares different sampling schemes.
We also analyze the effects of our design choices.
We defer details of implementation and hyper-parameters to \cref{apdx:experiment-details}.

\subsection{Datasets}

To demonstrate the usefulness in various usecases, we use five benchmark datasets with different
characteristics.
Here, we provide a brief overview of the datasets.
More dataset statistics are shown in \cref{apdx:dataset-stats}.

\textbf{BEOID}~\cite{Damen_BEOID_2014} is a dataset of egocentric activity videos,
    containing diverse activities ranging from cooking to work-outs.
We adopt the training-validation split from \citet{Ma_SF-Net_ECCV_2020}.

\textbf{GTEA}~\cite{Fathi_GTEA_2011} also consists egocentric videos but focuses fine-grained daily activities in a kitchen.
The median number of action instances per video is 18 in an about 60-second video.
This number is by far the largest among the datasets used in this paper.

\textbf{THUMOS\,'14}~\cite{Jiang_THUMOS14_2014} has significant variations in the lengths and the
number of occurrences of action instances.
Following the convention~\cite{Wang_UntrimmedNets_2017,Nguyen_STPN_2018}, we use the validation set
for training and the test set for evaluation.

\textbf{FineAction}~\cite{Liu_FineAction_TIP_2022} is a large scale dataset for fine-grained action
detection.
The fine-grained nature of action categories and the sparsity of action instances make this dataset
extremely challenging for action detection.

\textbf{ActivityNet 1.3}~\cite{Caba_ActivityNet_CVPR_2015} is a large-scale video dataset for action
recognition and detection of 200 diverse action categories.
The majority of videos in this dataset have only one action instance, and the duration of each
action instance is much longer than that of the other datasets.

\subsection{Evaluation Metrics}
As evaluation metrics, we report mean average precision (mAP) at various thresholds for temporal
intersection over union (IoU) (see \citet{Jiang_THUMOS14_2014} for the formal definition).
Following convention~\cite{Lee_LACP_ICCV_2021,Li_HAAN_ECCV_2022}, when calculating the average mAP (Avg mAP), we
average mAP's at the thresholds from 0.1 to 0.7 with a step 0.1 for BEOID, GTEA, and THUMOS\,'14,
and from 0.5 to 0.95 with a step 0.05 for FineAction and ActivityNet 1.3.
All the reported results of our method are the average of eight runs with different random seeds.

\subsection{Main Results}\label{sec:results}

\Cref{tab:gtea-beoid} provides the experimental results on BEOID and GTEA, demonstrating that our
method outperforms the point-level methods in terms of the average mAP.
The intervals of the AAPL labels are three seconds, 
which incurs less annotation costs than that
for the point-level labels, as shown in \cref{sec:annotation-time}.
Therefore, the results in \cref{tab:gtea-beoid} not just show the better accuracy of the proposed
method but also indicate the superiority of our approach regarding the trade-off between detection
performance and annotation time.

\begin{figure*}
    \centering
    \includegraphics[width=0.8\textwidth]{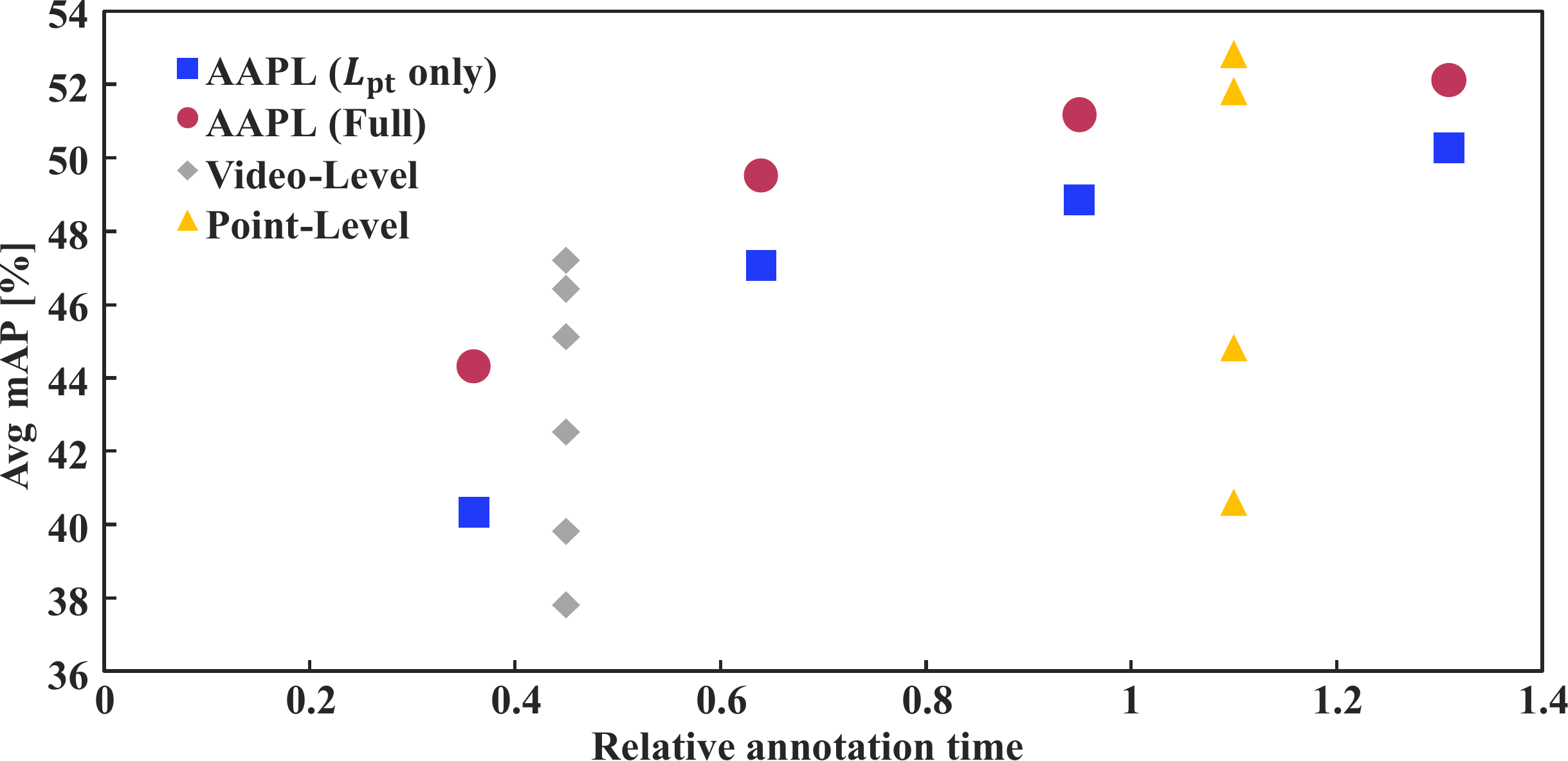}
    \caption{Trade-off between detection performance and annotation time.
    The blue squares represent AAPL-supervised training with $\losspt$ only,
    the red circles represent that with our full objective,
    the gray diamonds represent video-level methods~\cite{Paul_W-TALC_ECCV_2018,Min_A2CL-PT_ECCV_2020,Qu_ACM-Net_arXiv_2021,Huang_RSKP_CVPR_2022,Chen_DELU_ECCV_2022,Wang_AHLM_ICCV_2023},
    and the yellow triangles represent point-level methods~\cite{Ma_SF-Net_ECCV_2020,Ju_Point-Level_ICCV_2021,Lee_LACP_ICCV_2021,Li_PCL_ESWA_2023}.
    }
    \label{fig:thumos14-tradeoff}
\end{figure*}

\Cref{fig:thumos14-tradeoff} shows the trade-off between detection performance and annotation time
for AAPL-supervised learning on THUMOS\,'14,
including the results with our full objective and with $\losspt$ only.
The intervals between AAPL labels are 3, 5, 10, and 30 seconds, which are converted to the annotation times using \cref{tab:annotation-cost}.
For comparison, it also shows the results of previous one-stage training methods for
video-level~\cite{Paul_W-TALC_ECCV_2018,Min_A2CL-PT_ECCV_2020,Qu_ACM-Net_arXiv_2021,Huang_RSKP_CVPR_2022,Chen_DELU_ECCV_2022,Wang_AHLM_ICCV_2023}
and point-level~\cite{Ma_SF-Net_ECCV_2020,Ju_Point-Level_ICCV_2021,Lee_LACP_ICCV_2021,Li_PCL_ESWA_2023}
supervision.
For a fixed budget of annotation time, our full objective for AAPL
supervision is competitive with the state-of-the-art methods for the other types of supervision.
In addition, our baseline using $\losspt$ only already outperforms many of the previous methods.
Even the baseline with 30-second-interval AAPL labels achieves the average
    mAP comparable to that of \citet{Ma_SF-Net_ECCV_2020}, even though such
    sparse AAPL labels can be produced in one-third of the annotation time for point-level labels.
    This strength of the simple baseline illustrates the inherent effectiveness of AAPL supervision.

\begin{table*}[tb]
    \centering
    {
    \begin{tabular}{@{}llcccc@{}}
    \toprule
    & & \multicolumn{4}{c}{mAP@IoU\,[\%]}  \\ \cmidrule{3-6}
    Supervision & Method & 0.50 & 0.75 & 0.95 & Avg \\
    \midrule
    Video 
        & \citet{Paul_W-TALC_ECCV_2018}$^\dag$ & 6.2 & 3.2 & 0.8 & 3.5 \\
        & \citet{Lee_W-UM_AAAI_2021}$^\dag$ & 6.7 & 3.2 & 1.0 & 3.6 \\
        & \citet{Ma_ASL_CVPR_2021}$^\dag$ & 6.8 & 2.7 & 0.8 & 3.3 \\
        & \citet{Narayan_D2Net_ICCV_2021}$^\dag$ & 6.8 & 3.0 & 0.8 & 3.4 \\
        & \citet{Li_HAAN_ECCV_2022} & 7.1 & 4.0 & 1.1 & 4.1 \\
    \midrule
    Point &
    \citet{Lee_LACP_ICCV_2021}$^\ddag$ & 7.8 & 3.2 & 0.1 & 3.5 \\
    \midrule
    AAPL (30 sec.)& \emph{Ours} & 10.6 & 5.4 & 1.8 & 6.0 \\
    \bottomrule
    \end{tabular}
    }
    \caption{Comparison of detection performance on FineAction.
    $\dag$ indicates the results reproduced by \cite{Li_HAAN_ECCV_2022}, and 
    $\ddag$ indicates those reproduced by us (see \cref{apdx:lacp-fineaction} for details).}
    \label{tab:fineaction}
\end{table*}

The results on FineAction are shown in \Cref{tab:fineaction}.
It shows that even our proposed method with the most sparse labels outperforms all the previous methods,
demonstrating the strength of AAPL supervision with fine-grained and sparse actions.
Interestingly, LACP by \citet{Lee_LACP_ICCV_2021}, the point-level method outperforming all the video-level
methods on THUMOS\,'14, struggles with FineAction and falls short of the point-level approach, HAAN by \citet{Li_HAAN_ECCV_2022}.
We conjecture this is because of the sparsity of action instances in FineAction videos.
Very sparse point-level labels can enable the model to locate likely frames in actions but might not
be informative enough to help a detection model localize action boundaries.
This hypothesis is consistent with the fact that LACP outperforms HAAN in terms of
mAP@0.5, while lagging behind the latter at larger IoU thresholds; LACP successfully
found the actions but failed to localize them.

\begin{table*}[tb]
\centering
{
\begin{tabular}{@{}llllll@{}}
\toprule
& & \multicolumn{4}{c}{mAP@IoU\,[\%]} \\ \cmidrule{3-6}
Supervision & Method & 0.5 & 0.75 & 0.95 & Avg \\ \midrule
Point & \citet{Lee_LACP_ICCV_2021} & 40.4 & 24.6 & 5.7 & 25.1 \\ \midrule
AAPL (10 sec.) & \emph{Ours} & 41.3 & 25.4 & 5.8 & 25.7 \\
AAPL (30 sec.) & \emph{Ours} & 39.6 & 24.3 & 5.6 & 24.7 \\ \bottomrule
\end{tabular}
}
\caption{Detection performance on ActivityNet 1.3}
\label{tab:activitynet}
\end{table*}
\Cref{tab:activitynet} shows the results on ActivityNet 1.3.
For all the intervals we experimented with, our method achieved the detection accuracy
comparable or superior to \citet{Lee_LACP_ICCV_2021}, the state-of-the-art point-level method.

\subsection{Analysis}

In this section, we analyze and justify some of the design choices in our approach.

\subsubsection{Action-Agnostic Frame Sampling.} \label{sec:aafs}
\begin{table*}[tb]
\centering
{
\begin{tabular}{@{}lccccc@{}}
\toprule
& \multicolumn{4}{c}{THUMOS\,'14}
& BEOID \\
\cmidrule(lr){2-5} 
\cmidrule(lr){6-6}
&
    3 sec. & 5 sec. & 10 sec. & 30 sec. & 3 sec. \\
\midrule
Random sampling
    & 49.9
    & 49.9
    & 48.2
    & 43.8
    & 44.7 \\
Regular intervals
    & 52.1
    & 51.2
    & 49.5
    & 44.3
    & \textbf{55.2} \\
Clustering
    & \textbf{52.9}
    & \textbf{51.6}
    & \textbf{50.2}
    & \textbf{45.7}
    & 51.6 \\
\bottomrule
\end{tabular}
}
\caption{Comparison of different methods of action-agnostic frame sampling in terms of Avg mAP [\%].
The second row shows the average interval between AAPL labels.}
\label{tab:frame-sampling}
\end{table*}
The design of action-agnostic frame sampling impacts the detection performance.
To illustrate this, we conducted experiments with three different sampling schemes: random sampling,
regular intervals, and clustering-based sampling.
The clustering-based sampling first performs $k$-means clustering on snippet features extracted
using a pre-trained model and then selects the frames closest to the cluster centers (see
\cref{apdx:clustering} for details).
As shown in \cref{tab:frame-sampling}, the sampling at regular intervals and the clustering-based
sampling consistently outperform random sampling.

This suggests that annotating diverse frames is crucial for achieving good detection performance.
Indeed, both the regular-interval sampling and the clustering-based sampling tend to enhance the
diversity of the annotated frames:
the former does so by reducing the temporal correlation between the annotated frames, and the latter
by selecting frames that are separated in the embedding space.
Which of the two is better depends on the dataset, as shown in \cref{tab:frame-sampling}.

\subsubsection{Effectiveness of Each Component.}
\begin{table}[tb]
\centering
{
\begin{tabular}{@{}cccccc@{}}
\toprule
&&&
& \multicolumn{2}{c}{Avg mAP\,[\%]} \\
\cmidrule{5-6} 
$\losspt$ & $\lossvid$ & $\losspascl$ & PL &
    THUMOS\,'14 & BEOID
    \\
\midrule
\cmark &&&&
40.3 &
32.0 \\
\cmark & \cmark &&&
42.8 &
46.2 \\
\cmark & \cmark & \cmark &&
43.6 &
54.1 \\
\cmark & \cmark & \cmark & \cmark &
\textbf{44.3} &
\textbf{55.2} \\
\bottomrule
\end{tabular}
}
\caption{Effectiveness of each component. AAPL labels for THUMOS\,'14 are sampled at intervals of 30 seconds, and those for BEOID are sampled at intervals of 3 seconds.
``PL'' stands for ground-truth anchored pseudo-labeling.
}
\label{tab:ablation}
\end{table}
The proposed loss function consists of three components: $\losspt$, $\lossvid$, and $\losspascl$.
We also adopt the ground-truth anchored pseudo-labeling (PL) strategy.
To evaluate the effectiveness of each component, we conducted the ablation study, as shown in
\cref{tab:ablation}.
For both THUMOS\,'14 and BEOID, adding each component improves the detection accuracy, and the full
objective achieves the best performance.
The video loss makes particularly large contributions, showing that the self-training strategy based
on top-/bottom-$k$ pooling is effective with AAPL supervision, as with conventional weak supervision.

\subsubsection{Form of the Video Loss.}

\begin{table*}[tb]
\centering
{
\begin{tabular}{@{}lcccccccccc@{}}
\toprule
& \multicolumn{5}{c}{mAP@IoU\,[\%] (THUMOS\,'14)}
& \multicolumn{5}{c}{mAP@IoU\,[\%] (BEOID)} \\
\cmidrule(lr){2-6} 
\cmidrule(lr){7-11}
&
    0.1 & 0.3 & 0.5 & 0.7 & Avg &
    0.1 & 0.3 & 0.5 & 0.7 & Avg \\
\midrule
$\losspt$ only &
60.1 & 51.4 & 33.6 & 13.3 & 40.3 &
45.3 & 38.4 & 29.2 & 12.3 & 32.0 \\
\midrule
$\losspt+$ BCE loss &
61.4 & 52.1 & 34.0 & 12.6 & 40.8 &
56.8 & 50.6 & 38.1 & \textbf{19.2} & 42.0 \\
$\losspt+\lossvid$ (Ours) &
\textbf{64.3} & \textbf{54.6} & \textbf{35.2} & \textbf{14.0} & \textbf{42.8} &
\textbf{64.0} & \textbf{56.1} & \textbf{42.0} & 18.2 & \textbf{46.2} \\
\bottomrule
\end{tabular}
}
\caption{Comparison of video losses. AAPL labels for THUMOS\,'14 are sampled at intervals of 30 seconds, and those for BEOID are sampled at intervals of 3 seconds.}
\label{tab:video-loss-modification}
\end{table*}
The proposed video loss is adapted specifically for AAPL supervision to handle the incompleteness of
the video-level labels.
To demonstrate the effectiveness of our design of the video loss, we compare our proposed video loss
with the binary cross-entropy (BCE loss), which is the de facto standard in the
field~\cite{Lee_LACP_ICCV_2021,Li_PCL_ESWA_2023}.
As shown in \cref{tab:video-loss-modification}, the forms of the video loss impact the detection
performance.
In particular, mAP's at lower IoU thresholds are affected more than those at higher thresholds.
This is reasonable because the video loss, as a ranking-based pseudo-labeling strategy, does not
concern the accurate localization of action instances but does help mine unlabeled instances.

\section{Conclusion}
\label{sec:conclusion}

We proposed action-agnostic point-level (AAPL) supervision for temporal action detection to achieve a better trade-off between action detection performance and annotation costs.
We also proposed an action detection model and the training method to exploit AAPL-labeled data.
Extensive empirical investigation suggested that AAPL supervision was competitive with or outperformed previous supervision schemes for a wide range of action detection benchmarks in terms of the cost-performance trade-off.
Further analyses justified our design choices, such as frame sampling at regular intervals and the form of the video loss.

\section*{Acknowledgments}
We thank Ryoma Ouchi for his efforts and commitment at the early stage of this project.

\appendix
\section*{Appendix}

\begin{table*}[bt]
\centering
{
\begin{tabular}{@{}llllllll@{}}
\toprule
& & \multicolumn{2}{l}{Number of videos} & \multicolumn{2}{l}{Duration [sec]} & \multicolumn{2}{l}{Number per video}
    \\ \cmidrule(r){3-4} \cmidrule(r){5-6} \cmidrule(){7-8}
Dataset name & Classes & Training & Validation & Video & Action & Actions & Classes \\
\midrule
BEOID~\cite{Damen_BEOID_2014} & 34 & 46 & 12 & 45.3 & 1.2 & 6 & 3 \\
GTEA~\cite{Fathi_GTEA_2011} & 7 & 21 & 7 & 62.8 & 1.8 & 18 & 6 \\
THUMOS\,'14~\cite{Jiang_THUMOS14_2014} & 20 & 200 & 213 & 150.4 & 2.9 & 8 & 1 \\
FineAction~\cite{Liu_FineAction_TIP_2022} & 106 & 8,440 & 4,174 & 101.5 & 1.3 & 2 & 1 \\
ActivityNet 1.3~\cite{Caba_ActivityNet_CVPR_2015}& 200 & 10,024 & 4,926 & 114.3 & 26.8 & 1 & 1 \\
\bottomrule
\end{tabular}
}
\caption{Basic dataset statistics.
Here, the column ``Classes'' shows the number of action categories, 
``Number of videos'' shows the number of videos in the training and validation sets, 
``Duration'' shows the median duration of the training videos and action instances, and
``Number per video'' shows the median numbers of action instances and unique action classes in one training video.
}
\label{tab:dataset-stats}
\end{table*}

\section{Dataset Statistics}
\label{apdx:dataset-stats}

In \cref{tab:dataset-stats}
we provide the basic statistics of the datasets \cite{Damen_BEOID_2014,Fathi_GTEA_2011,Jiang_THUMOS14_2014,Liu_FineAction_TIP_2022,Caba_ActivityNet_CVPR_2015} used in the experiments  to facilitate readers' qualitative understanding of the
experiments and measurement.

Most videos in BEOID and GTEA contain action instances of multiple classes, while those in the other datasets typically have instances of only one class.
They also have many action instances in one minute of a video.
Because of these differences, the two datasets are highly challenging for the video-level setting.
Indeed, despite their popularity in the literature on point-level supervision, they have not been used to benchmark weakly supervised temporal action localization using video-level labels.

THUMOS\,'14 and ActivityNet 1.3 frequently appear in the action detection literature.
A video in THUMOS\,'14 typically has action instances of only one class, but the number of action instances is also prominent in some videos.
The dataset has significant variation in the duration and the number of action instances, which makes this a challenging benchmark.
By contrast, most videos in ActivityNet 1.3 have only one action instance, and each action instance has a long duration compared with the other datasets.
ActivityNet 1.3 stands out in terms of scale, with the number of videos being significantly larger than BEOID, GTEA, and THUMOS\,'14. This large-scale dataset provides a diverse range of action classes.

The FineAction dataset is the newest of the five datasets.
Its scale is comparable to ActivityNet 1.3, but the fine-grained nature of action classes and the relatively short duration and the sparsity action instances make it challenging even for point-level supervision, which has achieved superior performance on THUMOS\,'14 and ActivityNet 1.3 compared with video-level supervision.

\section{Details on Experiments\label{apdx:experiment-details}}

In this section, we provide the details of the experiments that are omitted in the main text.

\subsection{Implementation Details}
\label{sec:implementation-details}

In this section, we explain the details of the implementation of our method.

\subsubsection{Snippet Features.}

As the feature extractor, we employed the two-stream Inflated-3D (I3D) Inception-V1
model trained on Kinetics-400~\cite{Carreira_Kinetics_I3D_CVPR_2017}. It takes 16 frames
of RGB and optical flow, \ie, $N_S=16$, and outputs 1024-dimensional feature vectors for
each input modality. The snippet features were extracted before the experiments were
conducted, \ie, the feature extractor was frozen, and only the embedder and two heads
were updated during training. The two-stream features are concatenated into a single
snippet feature of 2048 dimensions, except for
FineAction~\cite{Liu_FineAction_TIP_2022}, whose two-stream feature has 4096 dimensions.
For GTEA and BEOID, we used I3D features extracted and distributed by
\cite{Ma_SF-Net_ECCV_2020}. For FineAction, we used I3D features made public by the
original dataset provider~\cite{Liu_FineAction_TIP_2022}. Following the previous
work~\cite{Li_HAAN_ECCV_2022}, we adopted the ``I3D\_100'' features, which the dataset
provider generated by temporally scaling the snippet feature sequence to the fixed
length of 100 by linear interpolation.

The FineAction paper~\cite{Liu_FineAction_TIP_2022} asserts that they have extracted
snippet features using ``I3D model'' with citation of the I3D
paper~\cite{Carreira_Kinetics_I3D_CVPR_2017}. The latter paper implemented the model as
the 3D version of Inception-V1~\cite{Szegedy_InceptionV1_2014}, which outputs 2048-dimensional features in the
two-stream setting. However, the distributed features of FineAction have 4096
dimensions. As of writing this paper, we have not figured out the exact setting in which
the features were extracted, but we used the distributed features as they were for a
fair comparison with previous work using the same features~\cite{Li_HAAN_ECCV_2022}.

The videos in a training dataset have varying durations, which result in varying lengths
of snippet feature sequences. To facilitate mini-batch training, we sampled the snippet
feature sequences to a fixed length. The sampled sequence lengths are 750 for
THUMOS\,'14, 50 for ActivityNet 1.3, and 100 for the other datasets.

\subsubsection{Scoring Model.}

As the feature embedder, we employed a single convolution layer with kernel size three,
followed by the ReLU activation function. The input and output of the embedder are the
same shape.

The classification and actionness heads accept half of the embedded features, \eg, in
the case of THUMOS\,'14 (or any dataset other than FineAction), the first 1024
dimensions of the embedded features are fed into the classification head, while the
others are consumed by the actionness head. Then, the heads apply the point-wise
convolution and the sigmoid activation function. During the training, the dropout is
also applied before the point-wise convolution. The dropout rate is set at 0.7 for all
the experiments.

\subsubsection{\texorpdfstring{Parametrization of the Top-$k$ Pooling}{Parametrization of the Top-k Pooling}.}

Following previous
work~\cite{Paul_W-TALC_ECCV_2018,Lee_BASNet_2020,Qu_ACM-Net_arXiv_2021}, we set $k_X$ in
the top-$k$ pooling as $k_X = \min \{ 1, \lfloor T / r_X \rfloor \}$, where $X$ is
either $\mathrm{fg}$ or $\mathrm{bg}$, $T$ is the length of the snippet feature
sequence, and $r_X$ is a hyper-parameter.
We borrowed the values of $r_\mathrm{fg}$ and $r_\mathrm{bg}$ from the previous
work~\cite{Qu_ACM-Net_arXiv_2021} and set $(r_\mathrm{fg}, r_\mathrm{bg})$ as $(8, 3)$
for the datasets other than ActivityNet~1.3 and $(2, 10)$ for ActivityNet~1.3.

\subsubsection{Optimization.}

The scoring model was optimized using the Adam optimizer with batch size 16. The
learning rate and weight decay were tuned by grid search.

\subsection{Hyper-Parameters}
\label{sec:hyper-params}

The hyper-parameters of the scoring model were tuned using the grid search. Because
optimizing many hyper-parameters is computationally expensive, we incrementally tuned
the hyper-parameters. Specifically, we first tuned the learning rate and weight decay,
then the weight of the video loss $\lambda\subvid$, then the weight of the contrastive
loss $\lambda\subpascl$ and the temperature $\tau$, and finally the thresholds
$\theta_\mathrm{fg}$ and $\theta_\mathrm{bg}$ for the ground-truth anchored
pseudo-labeling. From preliminary experiments, we found that the momentum parameter
$\mu$ of the prototype update for the contrastive loss, and therefore, we fixed it to
0.001 for all the experiments. The optimal values are summarized in
\cref{tab:hyper-params}.

\begin{table*}[tb]
\label{tab:hyper-params}
\centering
{
\begin{tabular}{@{}l|ll|l|ll|ll@{}}
\toprule
Dataset & LR & WD & $\lambda\subvid$ & $\lambda\subpascl$ & $\tau$ & $\theta_\mathrm{fg}$ & $\theta_\mathrm{bg}$ \\
\midrule
THUMOS\,'14 & $0.0001$ & $0.001$ & 3.0 & 0.1 & 0.10 & 1.0 & 0.05 \\
FineAction  & $0.0001$ & $0.0001$ & 0.003 & 0.01 & 0.1 & 1.0 & 0.5 \\
GTEA        & $0.0001$ & 0         & 0.3 & 0.03 & 0.3 & 0.5 & 0.0 \\
BEOID       & $0.001$ & $0.0001$ & 0.3 & 30.0 & 0.3 & 0.5 & 0.0 \\
ActivityNet 1.3 & $0.0001$ & $0.0001$ & 0.001 & 0.001 & 1 & 0.95 & 0.5 \\
\bottomrule
\end{tabular}
}
\caption{Optimal hyper-parameters for the scoring model. ``LR'' and ``WD'' are the
abbreviations of the learning rate and weight decay, respectively. The vertical lines
group the hyper-parameters that were tuned together.}
\end{table*}

\subsection{LACP Experiment on FineAction\label{apdx:lacp-fineaction}}

The evaluation of LACP~\cite{Lee_LACP_ICCV_2021} on FineAction was conducted using the
code developed by LACP's
authors\footnote{\url{https://github.com/Pilhyeon/Learning-Action-Completeness-from-Points/tree/1ef066a4d754b5a8c6993de5afc20898eea46118}}.
We modified the code to adapt it to the FineAction dataset and its evaluation metric.

Under the default configuration for THUMOS\,'14, the optimal sequence search
(OSS)~\cite{Lee_LACP_ICCV_2021} runs once in every ten iterations, but with the dataset
as large as FineAction, this frequency makes the training extremely inefficient.
Therefore, we dropped the frequency of OSS to once in every 200 iterations, which was 20
times as infrequent as the default value. However, this is twice as frequent in
terms of epochs as the default. We also swept the interval of 100 to 500 iterations
and verified that the 200-iteration interval achieved the peak detection accuracy.

Other hyper-parameters were tuned using the grid search. The optimal values that are
different from the default are the learning rate $2\times 10^{-5}$ and $r_\mathrm{act} = 2$.

\subsection{Clustering-Based Frame Sampling\label{apdx:clustering}}

Here, we describe the specific implementation of action-agnostic frame sampling based on clustering. 
The pseudo-code is presented in \cref{alg:algorithm}.
Basically, we perform the $k$-means clustering on the set of snippet features extracted by using pretrained feature extractor.
This algorithm is grounded on the assumption that it is more desirable to sample diverse frames.

We adopted the same feature extractor as the detection model, \ie, the Inflated 3D (I3D) convolutional neural network~\cite{Carreira_Kinetics_I3D_CVPR_2017} trained on the Kinetics dataset.
Feature vectors output by the I3D model have 2048 dimensions, but such high-dimensional features are not suitable for clustering.
For this reason, we apply the principal component analysis to reduce the dimensions to 64.
The number of clusters for a video is set to be $T^V/\tau$, where $T^V$ is the duration of the video $V$, and $\tau$ is the average interval of labeled frames.

\begin{algorithm}[tb]
\caption{Clustering-based frame sampling}
\label{alg:algorithm}
\begin{algorithmic}[1] 
\REQUIRE $V$: Video, $k$: The number of frames to sample, ${M}$: Pre-trained feature extractor, $D$: The number of principle components.
\ENSURE List of frames to pass to human annotators
\STATE Divide $V$ into a list of snippets $\mathcal{S}$.
\STATE Let $\mathcal{Z}$ be an empty list.
\FORALL{$S\in\mathcal{S}$}
    \STATE Append $M(S)$ to $\mathcal{Z}$.
\ENDFOR
\STATE Apply PCA on $\mathcal{Z}$ to get $\mathcal{Z}_\mathrm{PCA}$ with $D$ dimensions.
\STATE Apply $k$-means on $\mathcal{Z}_\mathrm{PCA}$.
\STATE Collect the cluster closest to the center of each cluster.
\RETURN the center frames from the collected snippets.
\end{algorithmic}
\end{algorithm}

\section{Measurement of Annotation Time\label{apdx:measurement}}

To demonstrate that the sparse supervision has benefits for annotation costs, we
measured actual annotation speed for three datasets,
THUMOS'14~\cite{Jiang_THUMOS14_2014}, GTEA~\cite{Fathi_GTEA_2011}, and
BEIOD~\cite{Damen_BEOID_2014}. This section explains the detailed protocol of the
measurement.

\subsection{Randomized Allocation}

The annotation time depends not only on the annotation method and the dataset but also
on various factors, such as an annotator's proficiency in the annotation method and
familiarity with the dataset. There can also be an issue of compatibility between videos
and annotation methods; some videos might be easier to annotate by one method than
others, and vice versa by another method. Ideally, it would be desirable to measure the
annotation time for the same set of videos using different annotation methods while
equalizing the effects of these factors. However, such an ideal measurement is
impossible because annotators become familiar with the annotation method and the dataset
as they proceed with the annotation. To overcome this problem, we employed the
randomized allocation strategy.

To this end, we hired eight workers and randomly allocated videos and annotation
methods. First, we arranged the videos in each dataset in a random order. Next, we
randomly assigned an annotation method (full, video-level, point-level, or AAPL with
different intervals) to each worker for each video. This allocation was carried out
using the block randomization
strategy~\cite{Fisher_ArrangementFieldExperiments_1926,RandomizedBlockDesigns}, which
ensures that each annotation method is assigned to each worker for the almost same
number of times. By this allocation, we aimed to extract the difference in annotation
time that is attributable to the annotation method while mitigating the effects of the
annotator's proficiency in the annotation methods, familiarity with the dataset, and the
compatibility issue between videos and the annotation methods.

Note that we reserved a few videos for the training of the workers. These videos were
selected so that at least one action instance per class appears in the videos. They were
used to familiarize the workers with the annotation methods and the action classes. The
reserved videos were not included in the measurement of the annotation time.

\subsection{Annotation Tool}

We developed a browser-based annotation tool for the four annotation methods. This tool
is the adaptation from VGG Image Annotator~\cite{Dutta_VIA_2019}, an open-source image
annotation tool equipped with the functionality for full supervision of temporal action
detection. The tool and the manual are available at \url{https://github.com/smy-nec/AAPL}.

\subsection{Instructions to Annotators}

In this section, we provide the instructions given to the annotators, except for the
usage of the tool, which is explained in the tool's manual. 

We instructed the workers to start the timer when they were ready to start annotating a
video, \ie, when they had finished reading the instructions, loading a VIA project file
and a video, and configuring the tool for annotation. We also instructed them to stop
the timer when they had reached the end of the video and finished the annotation. After
that, the workers were asked to save the project file containing the annotations and the
time measurement and to submit the project file to us. In addition, we asked the workers
to self-check the annotations. Self-checking for a video was performed immediately after
the annotation of the video was finished. The time of the self-checking was measured
separately from the first annotation.

Some workers might be so meticulous that they pause and repeat the video to ensure that
they produce as accurate annotations as possible. Although we acknowledge the importance
of keeping the annotations accurate, this kind of meticulousness can lead to an
overestimation of the annotation time. To mitigate this issue, we instruct the
workers not to rewind the video unless they need to do so because either
\begin{itemize}
    \item they have skipped a part of the video (\eg, for AAPL annotation) and need to
    go back a few seconds to decide the action occurring at that point, or
    \item they have passed an action boundary and need to go back to mark it,
\end{itemize}
and even if they did so, we asked them to rewind the video by a few seconds at most.

Regarding self-checking, we only asked them to check and correct clear mistakes such as
missing action instances and wrong action classes. We asked not to adjust the action
boundaries here because the self-checking is not intended to be a full re-annotation but
a quick check for quality control.

The workers were also given the textual definitions of each annotation method with
schematic figures like \cref{fig:supervision} in the main text. We also provided them with the list of
action classes, their definition, and examples of full annotations for the videos
reserved for the training of the workers.

\subsection{On the Measurement Result}

\begin{table*}[tb]
    \centering
    {
    \begin{tabular}{@{}lccccccc@{}}
    \toprule
    & Full & Video & Point & 3 sec & 5 sec & 10 sec & 30 sec \\
    \midrule
    THUMOS\,'14 & 2.994 & 0.810 & 1.863 & 2.272 & 1.648 & 1.072 & 0.644 \\
    GTEA & 6.105 & 1.591 & 4.594 & 3.138 & 2.481 & 1.690 & 0.855 \\
    BEOID & 5.205 & 1.976 & 3.873 & 3.305 & 2.312 & 1.483 & 0.827 \\
    \bottomrule
    \end{tabular}
    }
    \caption{Comparison of the relative annotation time, including the self-checking.
    Each cell shows the minutes it took for one annotator to annotate a 1-minute video.
    For AAPL supervision, we sampled frames to annotate at regular intervals of 3, 5,
    10, and 30 seconds.}
    \label{tbl:annotation-time-with-self-checking}
\end{table*}

The relative annotation time reported in the main text is the total annotation time divided by
the total annotated video duration.
The time measurement there does not include the time for the self-checking.
To complement those data, \cref{tbl:annotation-time-with-self-checking} shows the relative annotation time that also accounts for the self-checking.
This indicates that the comparison of annotation time among different annotation methods is not significantly affected by taking the self-checking into account.

\subsection{Limitation of the Current Measurement}

Actual annotation costs depend not only on the type of supervision but also on many details of an annotation pipeline, including the user interface of annotation software and instructions given to the workers.
Also, if the number of action categories is large as in ActivityNet and FineAction, searching a long list for the class of each action instance is unrealistic, and the design of the annotation pipeline must be fundamentally different from the one we implemented in this paper, as in \cite{Caba_ActivityNet_CVPR_2015}.
We would like to emphasize, however,
    that addressing the trade-offs between detection performance and annotation time is crucial in researching weak supervision schemes,
    and that measurement like ours is indispensable for this purpose.
Therefore, the annotation time as measured in this work should be considered not as the practically realistic value
    but as a convenient device to compare different supervision schemes on a reasonably equal footing.

\bibliographystyle{plainnat}
\bibliography{references}

\end{document}